\documentclass[sigconf]{acmart}
\usepackage{graphicx}
\usepackage{subcaption}
\usepackage{caption}
\usepackage{hyperref}
\usepackage{algorithm}
\usepackage{upgreek}
\usepackage{multirow}
\usepackage{algpseudocode}
\usepackage{amsmath}
\usepackage{natbib}
\usepackage{bm}
\usepackage{threeparttable}
\usepackage{makecell}
\usepackage{tabularx} % for better tables
\usepackage{booktabs} % for horizontal rules
\usepackage{array} 

\newcolumntype{M}[1]{>{\centering\arraybackslash}m{#1}}
\newcolumntype{C}[1]{>{\centering\arraybackslash}p{#1}}

\AtBeginDocument{%
  \providecommand\BibTeX{{%
    \normalfont B\kern-0.5em{\scshape i\kern-0.25em b}\kern-0.8em\TeX}}}

\setcopyright{acmcopyright}
\copyrightyear{2024}
\acmYear{2024}
\acmDOI{XXXXXXX.XXXXXXX}

\acmConference[DAC'24]{Make sure to enter the correct
  conference title from your rights confirmation email}{June 23--27,
  2024}{San Francisco, CA, USA}
\acmPrice{15.00}
\acmISBN{978-1-4503-XXXX-X/18/06}

\begin{document}

%%
%% The "title" command has an optional parameter,
%% allowing the author to define a "short title" to be used in page headers.
\title{Genetic Quantization-Aware Approximation for Non-Linear Operations in Transformers}
\author{Pingcheng Dong$^{1,2,}$\footnotemark[1], Yonghao Tan$^{1,2,}$\footnotemark[1], Dong Zhang$^{1,2}$, Tianwei Ni$^{3}$, Xuejiao Liu$^{2}$, Yu Liu$^{2}$, \\Peng Luo$^{2}$, Luhong Liang$^{2}$, Shih-Yang Liu$^{1}$, Xijie Huang$^{1}$, Huaiyu Zhu$^{3}$, \\Yun Pan$^{3}$, Fengwei An$^{4}$, Kwang-Ting Cheng$^{1,2}$\footnotemark[2] \\
% \large{$^1$The Hong Kong University of Science and Technology, Hong Kong SAR. \\
% $^2$AI Chip Center for Emerging Smart Systems (ACCESS), Hong Kong SAR. \\
% $^3$Zhejiang University, China. $^4$Southern University of Science and Technology, China.} \\
\large{$^1$The Hong Kong University of Science and Technology, Hong Kong SAR. $^2$AI Chip Center for Emerging Smart Systems (ACCESS), Hong Kong SAR. $^3$Zhejiang University, China. $^4$Southern University of Science and Technology, China.} \\
\large{\{pdongaa, ytanaz\}@connect.ust.hk, dongz@ust.hk, nitianwei@zju.edu.cn}, \\
\large{\{xuejiaoliu, albertliu, pengluo, luhong\}@ust.hk, \{sliuau, xhuangbs\}@connect.ust.hk}, \\
\large{\{zhuhuaiyu, panyun\}@zju.edu.cn, anfw@sustech.edu.cn, timcheng@ust.hk}
}
% \footnote{Both authors contributed equally to this research.}
% \authornote{Corresponding author.}

%
%%
%% By default, the full list of authors will be used in the page
%% headers. Often, this list is too long, and will overlap
%% other information printed in the page headers. This command allows
%% the author to define a more concise list
%% of authors' names for this purpose.
\renewcommand{\shortauthors}{Trovato and Tobin, et al.}

%%
%% The abstract is a short summary of the work to be presented in the
%% article.
\begin{abstract}
Non-linear functions are prevalent in Transformers and their lightweight variants, incurring substantial and frequently underestimated hardware costs. Previous state-of-the-art works optimize these operations by piece-wise linear approximation and store the parameters in look-up tables (LUT), but most of them require unfriendly high-precision arithmetics such as FP/INT 32 and lack consideration of integer-only INT quantization. This paper proposed a genetic LUT-Approximation algorithm namely \textit{GQA-LUT} that can automatically determine the parameters with quantization awareness. The results demonstrate that \textit{GQA-LUT} achieves negligible degradation on the challenging semantic segmentation task for both vanilla and linear Transformer models. Besides, proposed \textit{GQA-LUT} enables the employment of INT8-based LUT-Approximation that achieves an area savings of 81.3\textasciitilde81.7\% and a power reduction of 79.3\textasciitilde80.2\% compared to the high-precision FP/INT 32 alternatives. Code is available at  \textit{\url{https://github.com/PingchengDong/GQA-LUT}.}
\end{abstract}

%%
%% The code below is generated by the tool at http://dl.acm.org/ccs.cfm.
%% Please copy and paste the code instead of the example below.
%%
% \begin{CCSXML}
% <ccs2012>
%    <concept>
%        <concept_id>10010583.10010786.10010787.10010791</concept_id>
%        <concept_desc>Hardware~Emerging tools and methodologies</concept_desc>
%        <concept_significance>500</concept_significance>
%        </concept>
%    <concept>
%        <concept_id>10010520.10010521.10010542.10010294</concept_id>
%        <concept_desc>Computer systems organization~Neural networks</concept_desc>
%        <concept_significance>500</concept_significance>
%        </concept>
%  </ccs2012>
% \end{CCSXML}
% \ccsdesc[500]{Hardware~Emerging tools and methodologies}
% \ccsdesc[500]{Computer systems organization~Neural networks}

%%
%% Keywords. The author(s) should pick words that accurately describe
%% the work being presented. Separate the keywords with commas.
\keywords{Non-linear function, quantization-aware training, integer-only arithmetic, Transformer, look-up table, genetic algorithm \\
\textbf{ACM Reference Format:}\\
\small{Pingcheng Dong, Yonghao Tan, Dong Zhang, Tianwei Ni, Xuejiao Liu, Yu Liu, Peng Luo, Luhong Liang, Shih-Yang Liu, Xijie Huang, Huaiyu Zhu, Yun Pan, Fengwei An, Kwang-Ting Cheng. Genetic Quantization-Aware Approximation for Non-Linear Operations in Transformers. In \textit{Proceedings of DAC 2024: 61st IEEE/ACM Automation Conference. (DAC'24).} ACM, New York, NY, USA, 6 pages. \url{https://doi.org/XXXXX.XXXXX}}
\vspace{-1mm}
}

%% A "teaser" image appears between the author and affiliation
%% information and the body of the document, and typically spans the
%% page.
% \begin{teaserfigure}
%   \includegraphics[width=\textwidth]{sampleteaser}
%   \caption{Seattle Mariners at Spring Training, 2010.}
%   \Description{Enjoying the baseball game from the third-base
%   seats. Ichiro Suzuki preparing to bat.}
%   \label{fig:teaser}
% \end{teaserfigure}

% \received{20 February 2007}
% \received[revised]{12 March 2009}
% \received[accepted]{5 June 2009}

%%
%% This command processes the author and affiliation and title
%% information and builds the first part of the formatted document.
\maketitle
% \footnotetext[2]{First author's footnote.}
\section{Introduction}
\renewcommand{\thefootnote}{\fnsymbol{footnote}} % 改变脚注标记为符号
\footnotetext[1]{Both authors contributed equally.}
\renewcommand{\thefootnote}{\fnsymbol{footnote}} % 之后的脚注回到正常的数字编号
\footnotetext[2]{Corresponding author.}
The advent of the Transformer-based neural networks marked a new era for natural language processing \cite{kenton2019bert} and computer vision tasks \cite{liu2021swin, xie2021segformer}. The performance greatly benefits from the self-attention mechanism in Transformers, which could capture long-range dependencies well, but with a substantial overhead in computation and memory. Extensive research has been conducted to facilitate the deployment of  Transformers on edge devices. Techniques like lightweight structure integrating convolution and linear attention \cite{Cai_2023_ICCV, zhang2023augmented} emerge, while quantization \cite{kim2021bert,pmlr-v202-liu23w, liu2023llm} and run-time pruning \cite{tu2023multcim} has become favored approaches to further reduced the hardware burden. However, the optimization of non-linear operations is frequently neglected in Transformer-based models which can be costly due to the frequent involvement of high-precision operations like 32-bit floating-point (FP32) or integer (INT32) arithmetic. As reported by \cite{stevens2021softermax}, the inefficiency of non-linear operations seriously impedes the speed-up of Transformers at lots of hardware platforms.

Several prior works have endeavored to tackle the computational overheads posed by non-linear operations. Kim et al. \cite{kim2021bert} proposed to approximate the GELU, Softmax, and LayerNorm with quantization-aware 32-bit integer (INT32) arithmetic. Stevens et al. \cite{stevens2021softermax} customized a low-precision Softmax with a base replacement strategy. These methods can approximate and accelerate non-linear operations with minimal accuracy degradation but lack generality since each optimized operator contains distinct computational dataflow. To this end, a neural network-based general LUT-Approximation framework called \textit{NN-LUT} is designed by \cite{yu2022nn}. However, the LUT contains numerous parameters that are independent of the input data range. To utilize fewer hardware resources and take into account the range of input, the above methods are simplified in \cite{kim2023range} by factorizing floating-point data and performing single-entry LUT-Approximation on the mantissa in a small range. Although these excellent works progressively improve the generality and hardware-friendliness of the LUT-Approximation, they are still towards the computations in high-precision such as FP/INT32.

As for some customized neural network accelerators \cite{tu2023multcim}, the quantization scheme is generally in single-precision (e.g., INT8) or mixed-bit format \cite{huang2022sdq, 10071554}, and the inference may follow the dyadic arithmetic pipeline \cite{jacob2018quantization} to achieve integer-only computation. The I-BERT \cite{kim2021bert} is in the same category of integer-only quantization, but its management of the scaling factor and the input bit-width varies from operators. 
If high-precision LUT-Approximation is applied to INT8 inputs, it inevitably leads to resource wastage since the expressiveness of INT8 is much more limited compared to FP/INT32. Moreover, the method in \cite{kim2023range} cannot be utilized directly due to it is based on the floating-point decomposition strategy. The study from \cite{kim2023range} also highlights a notable variation in the representation range across layers. However, the \textit{NN-LUT} struggles to encapsulate this attribute unless subjected to calibration, a process that proves to be considerably time-consuming. This study begins with an in-depth analysis of the integer-only quantization scheme and LUT-Approximation. We identify the scaling factor in quantization as a crucial element that significantly impacts the precision of approximation parameters, noting that larger scaling factors lead to less accurate approximations. Building on this insight, we proposed a genetic quantization-aware algorithm to automatically determine the breakpoints in LUT-Approximation for non-linear functions. Termed as \textit{GQA-LUT}, this method paves the way for efficient hardware design with compact resource utilization, capitalizing on low-bit integer arithmetic. To effectively manage larger scaling factors, we further implement an algorithm that images the fixed-point (FXP) conversion as a form of mutation to boost the accuracy. The key contributions of our work are summarized as follows:
\vspace{-1mm}  % 减少下方空白
\begin{itemize}
\item We delve into the interplay between the scaling factor and LUT parameters, and formulate a general quantization-aware LUT-Approximation computing flow.
\item A genetic algorithm \textit{GQA-LUT} for automatic approximation is proposed, to overcome constraints of existing quantization algorithms that fail to adjust parameters by scaling factors.
\item A rounding mutation algorithm, which incorporates rounding error into the \textit{GQA-LUT}, is proposed to solve the breakpoint deviation issue while handling intractable scales
\item The area and power performance synthesized with TSMC 28-nm technology demonstrates that the INT8-based arithmetics achieve significant improvements, compared to the high-precision FP/INT32 units.
\end{itemize}

\begin{figure}[!t]
  \centering
  \begin{subfigure}{0.42\linewidth}
    \captionsetup{skip=1mm}
    \includegraphics[page=4, width=\linewidth]{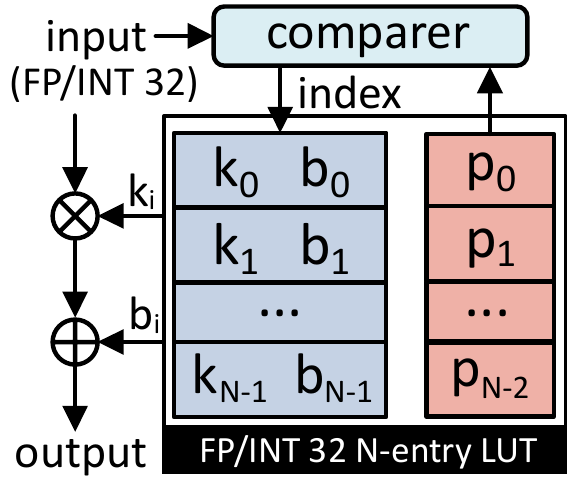}
    \caption{}
    \vspace{-4mm}  % 减少下方空白
    \label{Figure1(a)}
  \end{subfigure}
  \hfill % 添加一些水平空白，以便将子图分开
  \begin{subfigure}{0.55\linewidth}
    \captionsetup{skip=1mm}
    \includegraphics[page=5, width=\linewidth]{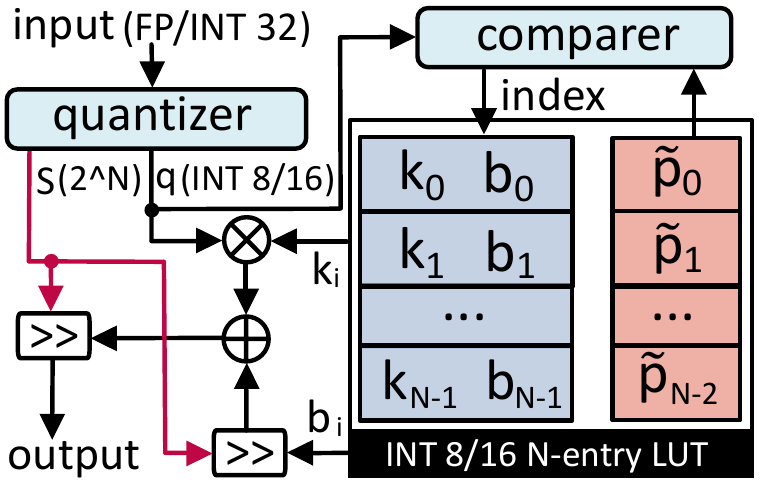}
    \caption{}
    \label{Figure1(b)}
    \vspace{-4mm}  % 减少下方空白
  \end{subfigure}
  \caption{Taxonomy of LUT-Approximation: (a) FP/INT32 LUT storage pattern, (b) INT8/16 LUT storage pattern with quantization awareness.}
  \label{Figure1}
  \vspace{-5mm}  % 减少下方空白
\end{figure}
\section{Preliminaries and Related Work}
\subsection{Non-Linear Operations in Transformers}
Non-linear operations are ubiquitous in vanilla transformer models \cite{vaswani2017attention}. The multi-head self-attention employs Softmax to capture the correlation in different dimensions, while GELU acts as the activation function within the feed-forward network (FFN), and LayerNorm is for the normalization. However, as the demand for edge computing increases, numerous lightweight transformer architectures have emerged, such as Softmax-free linear attention \cite{han2023flatten} and hybrid models that incorporate depthwise-separable convolution into early stages or FFNs \cite{Cai_2023_ICCV}. The non-linear operations in lightweight transformers are diverse such as HSWISH and cosine. Generally, these non-linear operators require high-precision computations with disparate dataflow to ensure accuracy. Hence, it becomes imperative to explore methods for handling diverse non-linear operations within a unified and simple hardware engine.
\subsection{LUT-based Approximation}
Adopting piece-wise linear approximation (\textit{pwl}) for various non-linear operations, coupled with parameter storage in LUT, has become a widely accepted strategy for enhancing hardware efficiency. This approach has been validated and endorsed by numerous recent and noteworthy studies \cite{yu2022nn, kim2023range}. All these LUT-based \textit{pwl} approximations can be formulated as follows:

\begin{equation}
pwl(x):=\left\{\begin{array}{cl}
k_0 x+b_0 & \text { if } x<p_0 \\
k_1 x+b_1 & \text { if } p_0 \leq x<p_1 \\
\vdots  \\
k_{N-1} x+b_{N-1} & \text { if } x \geqslant p_{N-2}
\end{array}\right.
\end{equation}
The $pwl(\cdot)$ function serves to approximate any nonlinear function $f(x)$ through a piece-wise linear approximation, comprising $N$ entries $\left\{pwl_i(x)\right\}_{i=0:N-1}$, each represented as a first-order function like $pwl_i(x)=k_ix+b_i$. The breakpoints $\left\{p_i\right\}_{i=0:N-2}$ determine the position of each entry. The parameters for an $N$-Entry $pwl$ are then stored in LUT as shown in Figure~\ref{Figure1(a)}. The accuracy could be readily maintained owing to the high representation ability of FP/INT32-based LUT storage and input data.  However, challenges may arise in scenarios involving low-bit quantization, a topic we delve into more in Section \ref{3}.

\subsection{Integer-Only Quantization}
Quantization is a popular technique to reduce the overhead of computation and memory after being deployed on chips since it can enable inference in a low-bit fashion. The quantization function is formulated as:
\begin{equation}
\tilde{x}=S \cdot q=S \cdot\left\lfloor\operatorname{Clip}\left(\frac{x}{S}, Q_n, Q_p\right)\right\rceil
\end{equation}
where $\lfloor\cdot\rceil$ represents the rounding function, $S$ is the scaling factor that bridges quantized values $q$ and high-precision (FP/INT32) inputs or weights $x$. In k-bit quantization, $q$ is obtained by clipping $\frac{x}{S}$ within $[-2^{k-1}, 2^{k-1}-1]$ for signed data or $[0, 2^{k}-1]$ for unsigned data, with the range defined by lower and upper bounds $Q_n$ and $Q_p$ respectively. The $q$ can be also dequantized to high-precision $\tilde{x}$ by multiplying with $S$. Commonly, $S$ is determined using the min-max method \cite{kim2021bert} or learned via the straight through estimator (STE) \cite{bengio2013estimating}, with the latter being more popular due to its proven effectiveness \cite{pmlr-v202-liu23w, DBLP:conf/iclr/EsserMBAM20}. In this work, we employ the LSQ \cite{DBLP:conf/iclr/EsserMBAM20} for quantization.

Integer-only quantization can be achieved by converting several scaling factors $S$ into dyadic numbers \cite{jacob2018quantization} or by excluding $S$ during inference \cite{kim2021bert}. However, the former method performs the non-linear function on the dequantized $\tilde{x}$ predominantly, which turns around to high-precision computation again. This stems from the fact that the non-linear function is linearly inseparable as discussed in \cite{kim2021bert}, e.g., $exp(S\cdot q)\neq S\cdot exp(q)$. The approximation in \cite{kim2021bert} can be directly applied on $q$ to achieve integer-only arithmetic, but lacks universality due to the diverse dataflow. As for the $pwl$ method, it can bring separability to the non-linear function by its inherence of $pwl(S\cdot q)= S\cdot pwl(q)$, whereas current $pwl$-based works \cite{yu2022nn, kim2023range} only focus on optimizing $pwl(S\cdot q)$, their arithmetics are still in FP/INT32 format. 
\section{Method}
\label{3}
\begin{algorithm}[!t]
  \caption{Genetic Piece-Wise Linear Approximation.}
  \label{alg:Framwork}
  \begin{algorithmic}[1]
    \Require
      Non-linear function $\bm{f(\cdot)}$, breakpoint size $\bm{N_b}$, population size $\bm{N_p}$, mutation function $\bm{M(\cdot)}$, cross probability $\bm{\theta_c}$, mutant probability $\bm{\theta_m}$ and $\bm{\theta_r}$, search range $\bm{[R_n, R_p]}$, iteration size $\bm{T}$, and the decimal bitwidth $\bm{\lambda}$ of slopes and intercepts.
    \Ensure
      Sets of slopes $\bm{\mathcal{K}}$, intercepts $\bm{\mathcal{B}}$ and breakpoints $\bm{\mathcal{P}}$
    \State Initialize a breakpoint population $\mathcal{O} = \{\bm{\mathcal{P}}_0, \bm{\mathcal{P}}_1, \ldots, \bm{\mathcal{P}}_{N_p-1}\}$, each $\bm{\mathcal{P}}_{i\in[0,N_p-1]}$ is a set of $\bm{N_b}$ random FP32 values in $\bm{[R_n, R_p]}$
    \For{$n = 0$ to $\bm{T}-1$} \Comment{$\bm{T}$-round evolution}
        % \For{$i = 0$ to $\bm{N_p}-1$} 
        \For{$\bm{\mathcal{P}}_i$ in $\mathcal{O}_n$} \Comment{$\mathcal{O}_n:$ The $n^{th}$ generation} 
        \State Initialize the Mean Squared Error (MSE): $E_i = 0$
        \State Create $pwl(\cdot)$ based on breakpoint set $\bm{\mathcal{P}}_i$
            \For{$x = R_n$ to $R_p$ with step $0.01$}
                \State Update $E_i = E_i + \frac{(pwl(x) - \bm{f}(x))^2}{(R_p - R_n)/0.01}$
            \EndFor
            \State $rand_c/rand_m \gets$ random number in $[0, 1]$
            \If{$rand_c < \bm{\theta_c}$}
                \State Randomly select $\bm{\mathcal{P}}_j$ from  $\mathcal{O} \setminus \{\bm{\mathcal{P}}_i\}$
                \State Swap a random segment between $\bm{\mathcal{P}}_j$ and $\bm{\mathcal{P}}_i$
            \EndIf
            \If{$rand_m < \bm{\theta_m}$}
                \State $\bm{M(\mathcal{P}_i, \bm{\theta_r})}$ // Perform mutation function $\bm{M(\cdot)}$
            \EndIf
        \EndFor
        \State $\mathcal{O}_{n+1} \gets$ Perform 3-size tournament selection on $\mathcal{O}_n$ 
    \EndFor
    \State $\bm{\mathcal{P}}=\bm{\mathcal{P}^*} \gets$ Select the best individual from $\mathcal{O}_T$
    \State $\bm{\mathcal{K}^*}, \bm{\mathcal{B}^*} \gets$ Derived from $\bm{\mathcal{P}^*}$
    % \State $\bm{\mathcal{P}} = \mathcal{P}^*$; $\bm{\mathcal{K}/\mathcal{B}} = \frac{\left\lfloor \bm{\mathcal{K}^*/\mathcal{B}^*} \cdot 2^{\bm{\lambda}} \right\rceil}{2^{\bm{\lambda}}}$ 
    \State $\bm{\mathcal{K}} = \frac{\left\lfloor \bm{\mathcal{K}^*} \cdot 2^{\bm{\lambda}} \right\rceil}{2^{\bm{\lambda}}}$, $\bm{\mathcal{B}} = \frac{\left\lfloor \bm{\mathcal{B}^*} \cdot 2^{\bm{\lambda}} \right\rceil}{2^{\bm{\lambda}}}$ \Comment{Round to FXP based on $\bm{\lambda}$}
  \end{algorithmic}
\end{algorithm} 
\subsection{Quantization-Aware Approximation}
\label{3.1}
Nothing that the $pwl$ affords us a unique opportunity to conduct $pwl$ directly on $q$ with the separation of scaling factor $S$. Instead of following the dyadic pipeline, we enforce the scaling factor to be the power-of-two by rounding the logarithmic value of a learnable parameter $\alpha$ to its nearest integer. Then, the scaling factor can be derived by $S=2^{\lfloor log_2^\alpha\rceil}$, and the STE is adopted to approximate the gradient in the non-differentiable round function. This power-of-two adaptation of $S$ is aimed at streamling the quantization-aware approximation, particularly for the parameters $\left\{b_i\right\}_{i=0:N-1}$ and $\left\{p_i\right\}_{i=0:N-2}$, expressed as:
\begin{equation}
\tilde{b}_i=\frac{b_i}{S}=b_i \gg \lfloor log_2^\alpha\rceil, \: \tilde{p}_i=\left\lfloor\operatorname{Clip}\left(\frac{p_i}{S}, Q_n, Q_p\right)\right\rceil
\label{eq:INTLUT}
\end{equation}
where the $\gg$ symbolizes the right shift operation, $\alpha$ represents the learnable parameter inherent to $S$, and $\tilde{b}_i$, $\tilde{p}_i$ denote the respective intercepts and breakpoints, scaled by the factor $S$ for $i$ ranging from $0$ to $N-1$ and $N-2$. Benefiting from the learnable power-of-two transformation, the low-cost shift operator can replace the divider while ensuring accuracy. Since the quantized input $q$ is an integer within $[Q_n, Q_p]$, the LUT only needs to store the original slopes $w_i$, intercepts $b_i$ and the quantized breakpoints $\tilde{p}_i$, where the $\tilde{b}_i$ is computed run-time by a shifter and the slopes are kept the same. Building upon this, we present a quantization-aware LUT-based approximation method in Figure~\ref{Figure1(b)}.

Nonetheless, specific non-linear operations like the divider (DIV) in Softmax and reciprocal of square root (RSQRT) in LayerNorm process intermediate fixed-point (FXP) outcomes directly, suspending the input quantization and consequently leading to extensive ranges of input data. To mitigate this, \textit{NN-LUT} implements input scaling, amplifying small values by multiplying them with a substantial constant. The \textit{RI-LUT} \cite{kim2023range} factorizes the floating-point values into a mantissa and a power-of-two scale. In essence, both methodologies can be categorized as quantization, with the scaling factor being chosen manually. In this work, we propose a \textit{Multi-Range Input Scaling} strategy that splits the input range outside the defined breakpoints interval $IR=[R_n, R_p]$ into $Ns$ distinct sub-ranges $SR_i=[SR_{ni}, SR_{pi}]$, where $0 \leq i < Ns-1$. The data in each sub-range is quantized to specific fixed-point values by a power-of-two scaling factor $S'_i$ chosen manually. The $pwl$ results will be multiplied with the  $S'_i$ and $\sqrt{S'_i}$ respectively for DIV and RSQRT to ensure the correct approximation. In contrast, the multi-range input scaling approach renders the scaling of breakpoints and intercepts unnecessary, as showcased in Equation (\ref{eq:INTLUT}), due to an intrinsic re-scaling process. Thus, the breakpoint and the intercepts could be rounded to a specific FXP format in this case.

\subsection{Genetic Piece-Wise Linear Approximation}
The performance of $pwl$ critically hinges on the judicious selection of optimal breakpoints. To address this, the \textit{NN-LUT} employs a novel neural network-based methodology. However, it necessitates substantial training data (100K) and presents challenges in managing the bitwidth of breakpoints during the training process. In this work, we propose to utilize a genetic algorithm namely \textit{GQA-LUT}, an optimization technique \cite{holland1992adaptation} inspired by the process of natural selection, to implement the $pwl$ as described in Algorithm \ref{alg:Framwork}. The main idea revolves around establishing a population containing several individuals which are the breakpoints $\bm{\mathcal{P}}$ of different $pwl(\cdot)$. The individuals in the population stochastically undergo crossover and mutation to enhance diversity and explore new solutions. The crossover swaps segments between pairs, while mutation introduces a normal distribution of noise. Subsequently, the individual exhibiting the highest fitness, which corresponds to the lowest mean squared error (MSE) in approximating the non-linear function $\bm{f(\cdot)}$, is identified and selected. The \textit{GQA-LUT} mirrors the principles of natural selection, where MSE serves as the criterion for selection.

\begin{figure*}[!t]
  \centering
    \setlength{\abovecaptionskip}{0.cm}
  \begin{subfigure}{0.5\linewidth}
    \captionsetup{skip=1mm}
    \includegraphics[width=\linewidth]{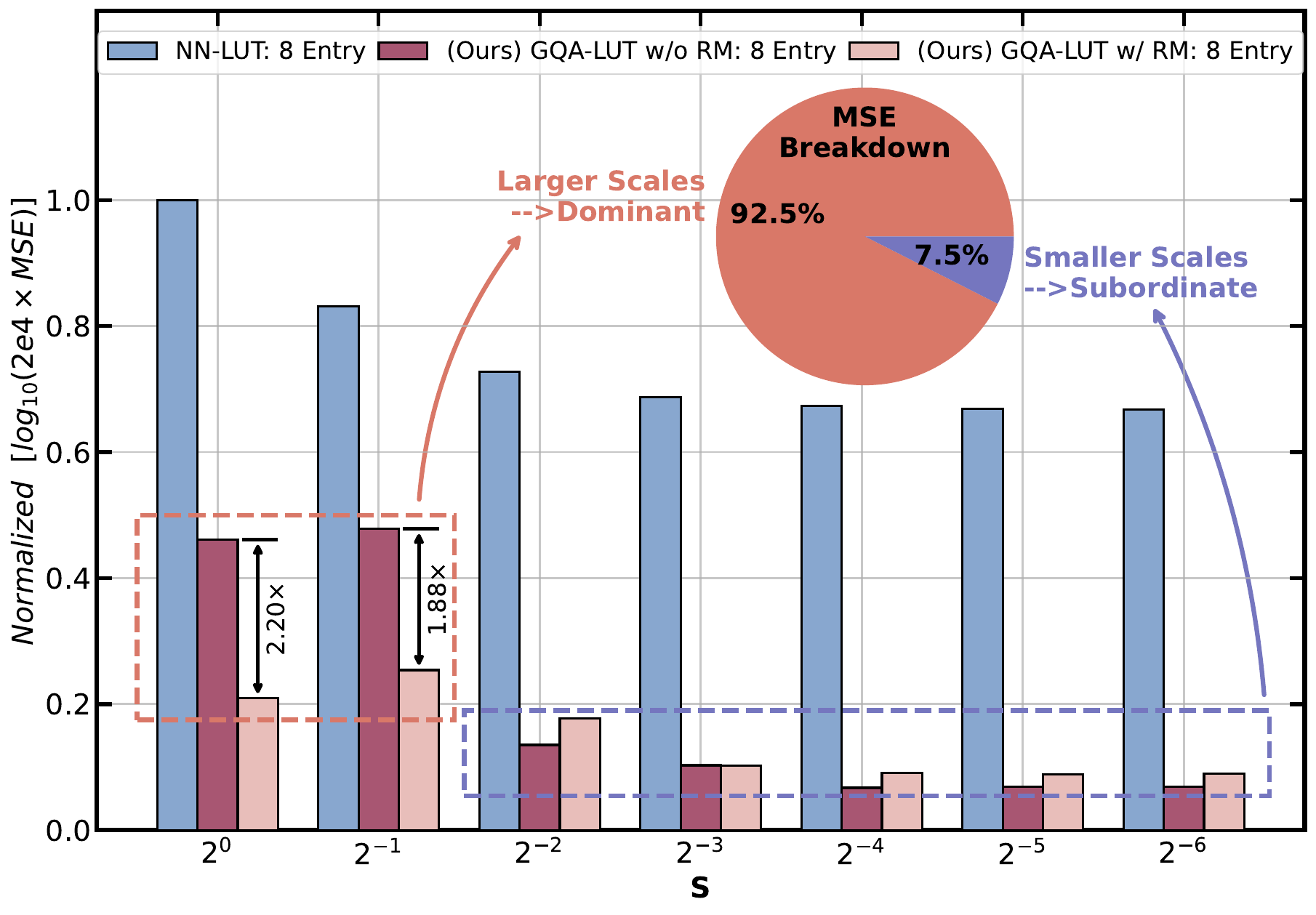}
    \caption{}
    \label{Figure2(a)}
  \end{subfigure}
  \hfill
  \begin{subfigure}{0.48\linewidth}
    \captionsetup{skip=1mm}
    \includegraphics[width=\linewidth]{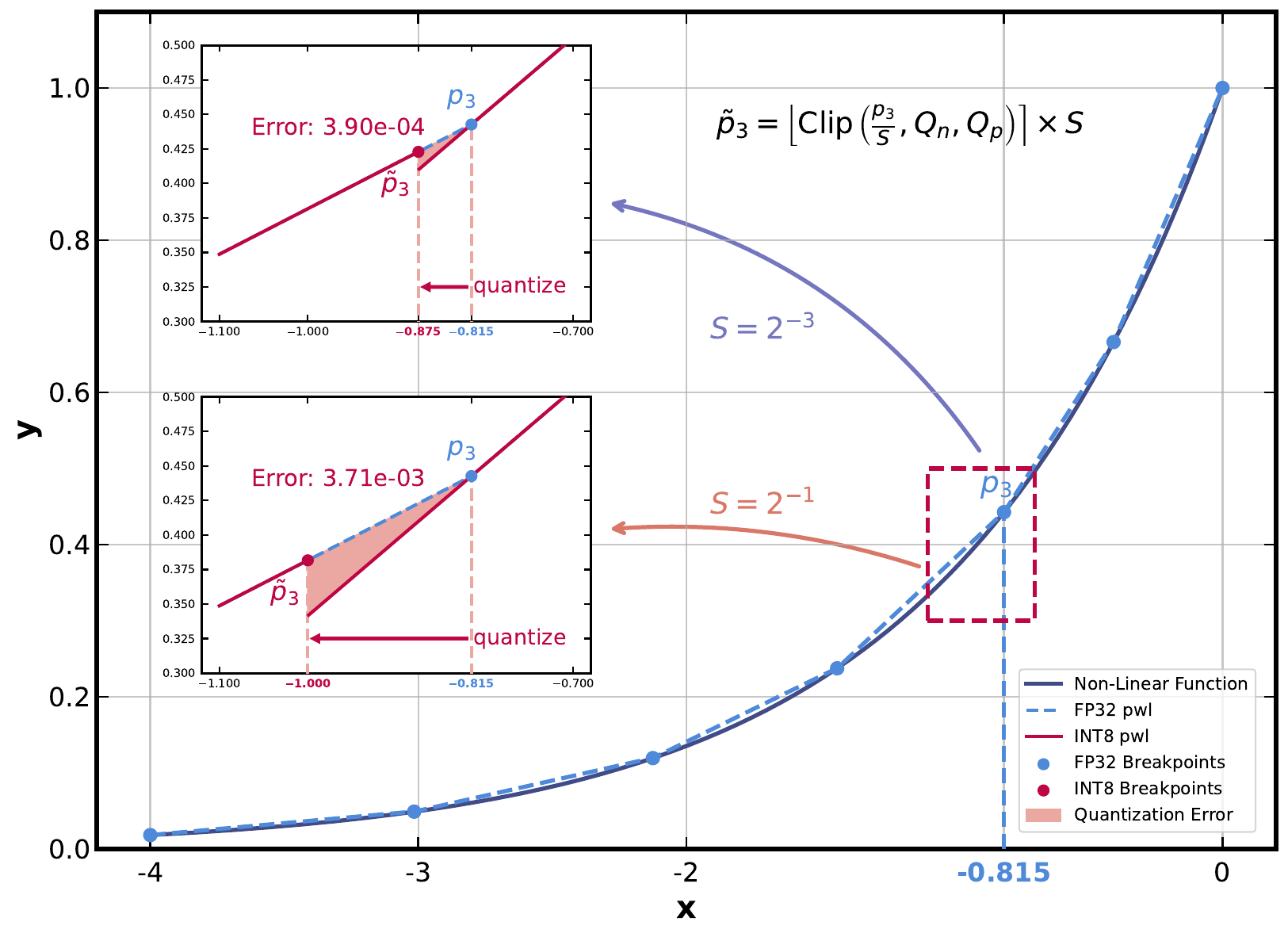}
    \caption{}
    \label{Figure2(b)}
  \end{subfigure}
  \caption{(a) Comparison of normalized MSE among \textit{NN-LUT}, \textit{GQA-LUT}, and \textit{GQA-LUT} with \textit{RM} strategy for GELU approximation using an 8-entry LUT, (b) breakpoint quantization analysis of GQA-LUT without \textit{RM} for EXP under different scaling factors.  }
  \label{Figure2}
\vspace{-1mm}  % 减少下方空白
\end{figure*}
To facilitate storage and computation in a low-bit INT format, we initially employ a straightforward approach, wherein the optimal sets of slopes and intercepts are converted from FP32 to FXP values based on a decimal bitwidth $\bm{\lambda}$. At the same time, the breakpoints are quantized as illustrated in Equation (\ref{eq:INTLUT}). The choice of $\bm{\lambda}$ is dictated by the search range $\bm{[R_n, R_p]}$ of a specific $\bm{f(\cdot)}$. We compare the accuracy of 8-entry \textit{GQA-LUT} and \textit{NN-LUT} under identical FXP conversion methods for GELU approximation in Figure \ref{Figure2(a)}. The results reveal that the \textit{GQA-LUT} exhibits superior accuracy to \textit{NN-LUT} across all scaling factors. Upon analyzing the MSE breakdown of \textit{GQA-LUT}, it becomes evident that the \textit{GQA-LUT} predominantly struggles with large $S$. Specifically, the scaling factors below $2^{-2}$ contribute to over 90 percent of the total error, highlighting a significant limitation in handling large $S$.
\begin{table}[!t]
\centering
\caption{Configurations of \textit{GQA-LUT} with \textit{RM} Strategy}
\label{table1}
\begin{tabular}{@{}C{1.6cm}|C{0.8cm}|C{1.2cm}|C{0.8cm}|C{0.8cm}|C{0.8cm}@{}} 
\hline
\hline
\multicolumn{1}{C{1.6cm}|}{\begin{tabular}[c]{@{}c@{}}Hyper- \\ parameters*\end{tabular}}  
& \multicolumn{1}{C{0.8cm}|}{GELU} 
& \multicolumn{1}{C{1.2cm}|}{HSWISH} 
& \multicolumn{1}{C{0.8cm}|}{EXP} 
& \multicolumn{1}{C{0.8cm}|}{DIV} 
& \multicolumn{1}{C{0.8cm}}{RSQRT} \\
\hline
\multicolumn{1}{C{1.6cm}|}{$[R_n, R_p]$}  
& \multicolumn{1}{C{0.8cm}|}{$(-4,4)$} 
& \multicolumn{1}{C{1.2cm}|}{$(-4,4)$} 
& \multicolumn{1}{C{0.8cm}|}{$(-8,0)$} 
& \multicolumn{1}{C{0.8cm}|}{$(0.5,4)$} 
& \multicolumn{1}{C{0.8cm}}{$(0.25,4)$} \\
\hline
\multicolumn{1}{C{1.6cm}|}{$\bm{\theta_r}$}  
& \multicolumn{1}{C{0.8cm}|}{0.05} 
& \multicolumn{1}{C{1.2cm}|}{0.05} 
& \multicolumn{1}{C{0.8cm}|}{0.05} 
& \multicolumn{1}{C{0.8cm}|}{0} 
& \multicolumn{1}{C{0.8cm}}{0} \\
\hline
\multicolumn{1}{C{1.6cm}|}{$[\bm{m_a}, \bm{m_b}]_8$}  
& \multicolumn{1}{C{0.8cm}|}{$[0,6]$} 
& \multicolumn{1}{C{1.2cm}|}{$[0,6]$} 
& \multicolumn{1}{C{0.8cm}|}{$[2,6]$}
& \multicolumn{1}{C{0.8cm}|}{-} 
& \multicolumn{1}{C{0.8cm}}{-} \\
\hline
\multicolumn{1}{C{1.6cm}|}{$[\bm{m_a}, \bm{m_b}]_{16}$} 
& \multicolumn{1}{C{0.8cm}|}{$[0,6]$}
& \multicolumn{1}{C{1.2cm}|}{$[2,6]$}
& \multicolumn{1}{C{0.8cm}|}{$[0,6]$} 
& \multicolumn{1}{C{0.8cm}|}{-} 
& \multicolumn{1}{C{0.8cm}}{-} \\
\hline
\multicolumn{1}{C{1.6cm}|}{Data Size}  
& \multicolumn{1}{C{0.8cm}|}{0.8K}
& \multicolumn{1}{C{1.2cm}|}{0.8K} 
& \multicolumn{1}{C{0.8cm}|}{0.8K} 
& \multicolumn{1}{C{0.8cm}|}{0.35K} 
& \multicolumn{1}{C{0.8cm}}{0.36K} \\
\hline
\hline
\multicolumn{6}{l}{\small*: $N_b=7$, $N_p=50$, $\theta_c=0.7$, $\theta_m=0.2$, $T=500$, and $\lambda=5$ by default.}
\vspace{-4mm}  
\end{tabular}
\end{table}
\begin{table}[!t]
\centering
\caption{Setup of \textit{Multi-Range Input Scaling} for Wide-Range \textit{DIV} and \textit{RSQRT} Operations under INT8 \textit{pwl}}
\label{table2}
\begin{tabular}{@{}C{0.8cm}|C{0.9cm}|C{1.3cm}|C{1.7cm}|C{2cm}@{}} 
\hline
\hline
\multicolumn{1}{C{0.8cm}|}{Ops*}
& \multicolumn{1}{C{0.9cm}|}{IR}
& \multicolumn{1}{C{1.3cm}|}{SR$_0/S'_0$}
& \multicolumn{1}{C{1.7cm}|}{SR$_1/S'_1$}
& \multicolumn{1}{C{2cm}}{SR$_2/S'_2$}  \\
\hline
\multicolumn{1}{C{0.8cm}|}{DIV}
& \multicolumn{1}{C{0.9cm}|}{$(0.5,4)$}
& \multicolumn{1}{C{1.3cm}|}{$[4,32)/2^{-3}$}
& \multicolumn{1}{C{1.7cm}|}{$[32,256)/2^{-6}$}
& \multicolumn{1}{C{2cm}}{$[256,+\infty)/2^{-6}$}\\
\hline
\multicolumn{1}{C{0.8cm}|}{RSQRT}
& \multicolumn{1}{C{0.9cm}|}{$(0.25,4)$}
& \multicolumn{1}{C{1.3cm}|}{$[4,64)/2^{-4}$}
& \multicolumn{1}{C{1.7cm}|}{$[64,1024)/2^{-8}$}
& \multicolumn{1}{C{2cm}}{$[1024,+\infty)/2^{-12}$}\\
\hline
\hline
\multicolumn{5}{l}{\small*: The breakpoints are rounded to 8bit FXP with $\lambda$ decimal bits.}
\end{tabular}
\vspace{-5mm}  
\end{table}
\subsection{Rounding Mutation}
To investigate why approximation errors predominate at larger values of $S$, we closely examine the approximation curves of the exponential function (EXP) depicted in Figure \ref{Figure2(b)}. When a breakpoint $p$ undergoes quantization to a specific integer value according to Equation (\ref{eq:INTLUT}), the resulting approximation error demonstrates variability across different scaling factors $S$. In particular, at larger $S$, the breakpoint is susceptible to significant shifts, leading to noticeable approximation offsets—a phenomenon we term as \textbf{breakpoint deviation}. In contrast, a smaller $S$ tends to yield a minimal deviation, thereby mitigating the error. This observation highlights the limitation of applying a straightforward FXP conversion to \textit{GQA-LUT}, especially at larger scaling factors, where breakpoint deviation becomes substantially more prominent.
\begin{algorithm}[!t]
  \caption{Rounding Mutation (RM) Algorithm.}
  \label{alg:rounding_mutation}
  \begin{algorithmic}[1]
    \Require
      Breakpoint set $\bm{\mathcal{P}}$, entry size $\bm{e}$, mutate range $\bm{[m_a, m_b]_{e}}$, and mutation probability $\bm{\theta_r}$.
    \Ensure
      Mutated breakpoint set $\bm{\mathcal{\Hat{P}}}$
    \State $\bm{\mathcal{\Hat{P}}} \gets \emptyset$ \Comment{Initialize the mutated set}
    % \State $\bm{\theta_r} \gets [0] + \bm{\theta_r}$ \Comment{Prepend $0$ to the mutation probability set}
    \For{each $p$ in $\bm{\mathcal{P}}$} \Comment{Mutate all individuals}
        \State $rand_p \gets \text{random number in } [0, 1]$ 
        \For{$i = \bm{m_a}$ to $\bm{m_b}$} \Comment{$0 \leq \bm{m_a} \leq \bm{m_b}$}
            \If{$i\cdot\bm{\theta_r} \leq rand_p < (i+1)\cdot\bm{\theta_r}$}
                \State $p' \gets \lfloor p \cdot 2^{i} \rceil / 2^{i}$ \Comment{Rounding mutation}
                \State Append $p'$ to $\bm{\mathcal{\Hat{P}}}$ 
                \State \textbf{break} \Comment{Mutate only once}
            \EndIf
        \EndFor
    \EndFor
    \State Sort $\bm{\mathcal{\Hat{P}}}$ in ascending order \Comment{Ensure correct order}
  \end{algorithmic}
\end{algorithm}
To tackle this challenge, we propose a novel strategy called Rounding Mutation (\textit{RM}), detailed in Algorithm \ref{alg:rounding_mutation}. The \textit{RM} method images the quantization of breakpoints as a stochastic mutation, wherein quantization is applied randomly across various scales $S$, influencing each element of an individual set of breakpoints throughout the evolutionary process. In \textit{GQA-LUT}, we substitute the conventional mutation function, which relies on normally distributed noise, with our \textit{RM} approach, all the while preserving the straightforward FXP conversion for both slopes and intercepts. As depicted in Figure \ref{Figure2(a)}, the integration of the \textit{RM} strategy within \textit{GQA-LUT} markedly diminishes the MSE when S is large. Though there is a minor escalation in MSE for smaller S that is originally subordinate, the increment is minimal and can be practically disregarded. On the other hand, incorporating \textit{RM} into \textit{NN-LUT} is intricate, as it relies on a neural network-based framework where breakpoints are deduced from the slopes and intercepts, a methodology that is inherently inverse to that of \textit{GQA-LUT}. The latter determines the slopes and intercepts directly from the breakpoints with varying precision levels. 

\begin{table}[!t]
\centering
\caption{Comparison of Average MSE on Different Methods}
\label{table3}
\begin{tabular}{@{}M{1.1cm}|C{0.6cm}|C{0.7cm}|C{1.2cm}|C{0.8cm}|C{0.8cm}|C{0.7cm}@{}} 
\hline
\hline
\multicolumn{1}{C{1.1cm}|}{Methods*}
& \multicolumn{1}{C{0.6cm}|}{Entry}
& \multicolumn{1}{C{0.7cm}|}{GELU}
& \multicolumn{1}{C{1.2cm}|}{HSWISH}
& \multicolumn{1}{C{0.8cm}|}{EXP}
& \multicolumn{1}{C{0.8cm}|}{DIV}
& \multicolumn{1}{C{0.7cm}}{RSQRT} \\
\hline
\multirow{2}{1.1cm}{\centering\small\textit{NN-LUT}} 
& 8 
& $1.3e^{-3}$ 
& $1.2e^{-3}$ 
& $6.4e^{-4}$ 
& $2.7e^{-3}$ 
& $1.1e^{-2}$  \\ 
\cline{2-7} 
& 16 
& $2.7e^{-4}$
& $7.9e^{-4}$ 
& $2.3e^{-4}$ 
& $2.4e^{-3}$ 
& $2.8e^{-3}$ \\ 
\hline
\multirow{2}{1.1cm}{\parbox{1.2cm}{\centering\small\textit{GQA-LUT}\\w/o \textit{RM}}}  
& 8 
& $1.5e^{-4}$ 
& $3.1e^{-4}$ 
& $1.3e^{-4}$ 
& $\bm{7.8e^{-4}}$ 
& $\bm{1.2e^{-3}}$  \\ 
\cline{2-7} 
& 16 
& $9.9e^{-5}$ 
& $2.8e^{-4}$ 
& $1.1e^{-4}$ 
& $1.3e^{-3}$ 
& $5.0e^{-4}$ \\ 
\hline
\multirow{2}{1.1cm}{\parbox{1.2cm}{\centering\small\textit{GQA-LUT}\\w/\textit{RM}}}  
& 8 
& $\bm{9.4e^{-5}}$ 
& $\bm{2.9e^{-4}}$
& $\bm{1.2e^{-4}}$ 
& $8.3e^{-4}$ 
& $1.7e^{-3}$  \\ 
\cline{2-7} 
& 16 
& $9.6e^{-5}$ 
& $2.2e^{-4}$ 
& $7.4e^{-5}$ 
& $1.4e^{-3}$ 
& $1.2e^{-3}$ \\ 
\hline
\hline
\multicolumn{7}{l}{\small*: All methods focus on INT8 LUT approximation.}
\end{tabular}
  \vspace{-5mm}  
\end{table}
\begin{table}[!t]
\centering
\caption{Fine-tuning mIoU of Segformer-B0 on Cityscapes}
\label{table4}
\begin{tabular}{@{}C{2cm}|C{1.5cm}|C{1.5cm}|C{1.5cm}@{}} 
\hline
\hline
\multicolumn{1}{C{2cm}|}{Replacement*}
& \multicolumn{1}{C{1.5cm}|}{\textit{NN-LUT}}
& \multicolumn{1}{C{1.5cm}|}{\begin{tabular}[c]{@{}c@{}} \textit{GQA-LUT} \\ \textit{w/o RM}\end{tabular}}
& \multicolumn{1}{C{1.5cm}}{\begin{tabular}[c]{@{}c@{}} \textit{GQA-LUT} \\ \textit{w/RM}\end{tabular}} \\
\hline
\multicolumn{1}{C{2cm}|}{None}
& \multicolumn{1}{C{1.5cm}|}{74.60\%}
& \multicolumn{1}{C{1.5cm}|}{74.60\%}
& \multicolumn{1}{C{1.5cm}}{74.60\%} \\
\hline
\multicolumn{1}{C{2cm}|}{EXP only}
& \multicolumn{1}{C{1.5cm}|}{73.88\%}
& \multicolumn{1}{C{1.5cm}|}{74.31\%}
& \multicolumn{1}{C{1.5cm}}{74.59\%} \\
\hline
\multicolumn{1}{C{2cm}|}{GELU only}  
& \multicolumn{1}{C{1.5cm}|}{73.61\%} 
& \multicolumn{1}{C{1.5cm}|}{74.24\%} 
& \multicolumn{1}{C{1.5cm}}{74.57\%} \\
\hline
\multicolumn{1}{C{2cm}|}{DIV only}  
& \multicolumn{1}{C{1.5cm}|}{74.15\%} 
& \multicolumn{1}{C{1.5cm}|}{74.37\%} 
& \multicolumn{1}{C{1.5cm}}{74.58\%}  \\
\hline
\multicolumn{1}{C{2cm}|}{RSQRT only}  
& \multicolumn{1}{C{1.5cm}|}{73.94\%} 
& \multicolumn{1}{C{1.5cm}|}{74.17\%} 
& \multicolumn{1}{C{1.5cm}}{74.51\%} \\
\hline
\multicolumn{1}{C{2cm}|}{Altogether}  
& \multicolumn{1}{C{1.5cm}|}{73.46\%$_{\color{red}{-1.14}}$} 
& \multicolumn{1}{C{1.5cm}|}{74.28\%$_{\color{red}{-0.32}}$} 
& \multicolumn{1}{C{1.5cm}}{74.53\%$_{\color{red}{-0.07}}$} \\
\hline
\hline
\multicolumn{4}{l}{\small*: The non-linear function that is replaced by 8-Entry $pwl$.}
\end{tabular}
  \vspace{-6mm}  
\end{table}

\section{Experimental Results}
In this section, a comprehensive comparison between \textit{GQA-LUT} and \textit{NN-LUT} is presented, initially focusing on operator-level accuracy for a variety of non-linear functions.  Following this, the fine-tuning accuracy of two models, Segformer \cite{xie2021segformer} and EfficientViT \cite{Cai_2023_ICCV}, specifically in the context of semantic segmentation tasks under integer-only quantization is evaluated. We further implement LUT-based $pwl$ with various precision using Verilog HDL and benchmark their hardware performances including area and power dissipation.

\subsection{Operator-Level Accuracy}
\begin{figure*}[!t]
  \centering
    \setlength{\abovecaptionskip}{0.cm}
  \includegraphics[width=\linewidth]{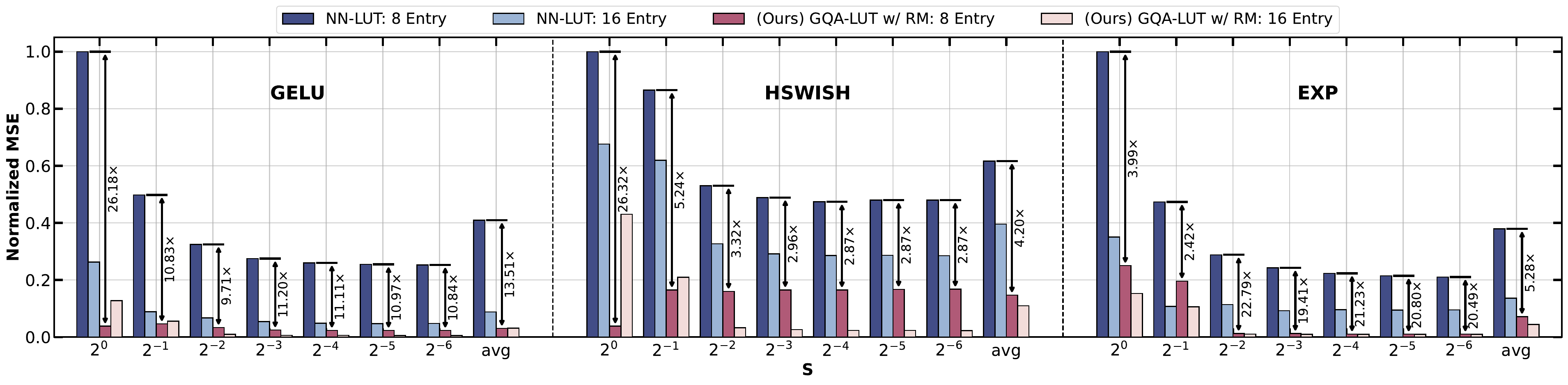}
  \caption{Comparison of normalized MSE for GELU, HSWISH, and EXP across various INT8 quantization scaling factors $S$ using \textit{NN-LUT}, \textit{GQA-LUT}, and the \textit{RM} strategy with 8/16-entry LUT approximation.}
  \label{Figure3}
  \vspace{-4mm}  
\end{figure*}

We investigate five common non-linear functions in the Transformer and its variants, specifically: GELU, EXP, HSWISH, DIV, and the RSQRT. The hyperparameters associated with each function, as implemented under \textit{GQA-LUT} are listed in Table \ref{table1}. As for wide-range operators DIV and RSQRT, the multi-range input scaling strategy is adopted whose setup is illustrated in Table \ref{table2}. We also re-implement \textit{NN-LUT} following the training procedure described in \cite{yu2022nn} and directly convert the slopes, intercepts, and breakpoints to the same precision as \textit{GQA-LUT}. To comprehensively evaluate operator-level accuracy with quantization awareness, we emphasize dequantized data instead of randomly selecting input data in the floating-point range \cite{yu2022nn}. During the evaluation, input data is orderly sampled from the dequantized range, for instance, $[Q_nS, Q_pS]$ with an incremental step size of S. Given that only GELU, EXP, and HSWISH are affected by the scaling factor $S$, we present a detailed MSE comparison for various $S$ values in Figure (\ref{Figure3}). It can be seen that the \textit{GQA-LUT w/RM} achieves a more stable and better performance across different $S$ than \textit{NN-LUT}. Furthermore, the accuracy for all operators presented in Table \ref{table3} distinctly illustrate the superior performance of \textit{GQA-LUT w/RM} for GELU, EXP, and HSWISH that containing input scaling factor, outshining the \textit{NN-LUT} even though both adopt INT8 integer-only quantization. But DIV and RSQRT that receive merely quantized input are apt to \textit{GQA-LUT w/o RM} which also further proves the \textit{RM} is born to handle data with changeful precision. Impressively, the required data of \textit{GQA-LUT w/RM} (0.35\textasciitilde0.8K) is extremely lower than that of the \textit{NN-LUT} (100K, reported in \cite{yu2022nn}).

\subsection{Fine-tuning Accuracy}
\begin{table}[!t]
\centering
\caption{Fine-tuning mIoU of EfficientViT-B0 on Cityscapes}
\label{table5}
\begin{tabular}{@{}C{2cm}|C{1.5cm}|C{1.5cm}|C{1.5cm}@{}} 
\hline
\hline
\multicolumn{1}{C{2cm}|}{Replacement*}
& \multicolumn{1}{C{1.5cm}|}{\textit{NN-LUT}}
& \multicolumn{1}{C{1.5cm}|}{\begin{tabular}[c]{@{}c@{}} \textit{GQA-LUT} \\ \textit{w/o RM}\end{tabular}}
& \multicolumn{1}{C{1.5cm}}{\begin{tabular}[c]{@{}c@{}} \textit{GQA-LUT} \\ \textit{w/RM}\end{tabular}} \\
\hline
\multicolumn{1}{C{2cm}|}{None}
& \multicolumn{1}{C{1.5cm}|}{74.17\%}
& \multicolumn{1}{C{1.5cm}|}{74.17\%}
& \multicolumn{1}{C{1.5cm}}{74.17\%} \\
\hline
\multicolumn{1}{C{2cm}|}{HSWISH only}
& \multicolumn{1}{C{1.5cm}|}{73.55\%}
& \multicolumn{1}{C{1.5cm}|}{73.98\%}
& \multicolumn{1}{C{1.5cm}}{74.20\%} \\
\hline
\multicolumn{1}{C{2cm}|}{DIV only}  
& \multicolumn{1}{C{1.5cm}|}{73.21\%} 
& \multicolumn{1}{C{1.5cm}|}{73.30\%} 
& \multicolumn{1}{C{1.5cm}}{74.08\%} \\
\hline
\multicolumn{1}{C{2cm}|}{Altogether}  
& \multicolumn{1}{C{1.5cm}|}{73.27\%$_{\color{red}{-0.90}}$} 
& \multicolumn{1}{C{1.5cm}|}{73.79\%$_{\color{red}{-0.38}}$} 
& \multicolumn{1}{C{1.5cm}}{74.15\%$_{\color{red}{-0.02}}$} \\
\hline
\hline
\multicolumn{4}{l}{\small*: The non-linear function that is replaced by 8-Entry $pwl$.}
\end{tabular}
  \vspace{-6mm}  
\end{table}

\begin{table}[!t]
\centering
\caption{Hardware Costs under TSMC 28-nm Technology}
\label{table6}
\begin{tabular}{@{}M{1.2cm}|C{1cm}|C{1.5cm}|C{1.6cm}@{}} 
\hline
\hline
\multicolumn{1}{C{1.2cm}|}{Precision*}
& \multicolumn{1}{C{1cm}|}{Entry}
& \multicolumn{1}{C{1.5cm}|}{Area (um$^2$)}
& \multicolumn{1}{C{1.6cm}}{Power (mW)} \\
\hline
\multirow{2}{1.2cm}{\centering\small\textit{INT8}} & 8 & 961 & 0.40 \\ 
\cline{2-4} 
 & 16 & 1640& 0.78 \\ 
\hline
\multirow{2}{1.2cm}{\centering\small\textit{INT16}}  & 8 & 2080 & 0.85   \\ 
\cline{2-4} 
 & 16 & 3521 & 1.47 \\ 
\hline
\multirow{2}{1.2cm}{\centering\small\textit{INT32}}  & 8 & 5243 & 1.93  \\ 
\cline{2-4} 
 & 16 & 8040 & 3.14 \\ 
\hline
\multirow{2}{1.2cm}{\centering\small\textit{FP32}}  & 8 & 5135 & 2.02  \\ 
\cline{2-4} 
 & 16 & 7913 & 3.47 \\ 
\hline
\hline
\multicolumn{4}{l}{\small*: Denotes the precision of input and LUT parameters.}
\end{tabular}
  \vspace{-6mm}  
\end{table}
We extend our research by examining fine-tuning accuracy on the challenging and demanding Cityscapes dataset, a benchmark in semantic segmentation \cite{cordts2016cityscapes}. The Cityscapes is annotated at the pixel level, depicting urban scenes across 19 varied categories. It comprises 2,975 images for training, 500 for validation, and 1,525 for testing, all showcased in a high-resolution $(1024 \times 2048)$ format. In our analysis, we focus on two specific Transformer models: the Segformer-B0 at $1024\times1024$ resolution, a vanilla Transformer incorporating EXP, GELU, DIV, and RSQRT as non-linear operators, and the EfficientViT-B0 at $1920\times1024$ resolution, a lightweight Transformer variant proposed recently, it only contains HSWISH and DIV operators based on linear attention for edge devices. To begin with, we apply INT8 integer-only quantization to both the weights and activations of the aforementioned models, utilizing the LSQ method \cite{DBLP:conf/iclr/EsserMBAM20} and adhering to the dyadic pipeline \cite{jacob2018quantization}. In contrast to the approach in I-BERT, which involves updating the scaling factor in the non-linear function, we restrict the scaling factor for the input of the non-linear function as a power-of-two format, in alignment with the methodology outlined in Section \ref{3.1}. These quantized models subsequently serve as our baseline for comparison. Following the existing methods \cite{Cai_2023_ICCV, xie2021segformer, zhang2022graph}, we employ the standard mean Intersection over Union (mIoU) as our primary metric for evaluating fine-tuning performance. The accuracy of fine-tuning Segformer-B0 and EfficientViT-B0 is presented in Table \ref{table4} and Table \ref{table5}, respectively. The results highlight that the \textit{GQA-LUT w/RM} effectively maintains the fine-tuning accuracy for both Segformer-B0 and EfficientViT-B0, with minimal loss of $0.07\%$ and $0.02\%$ respectively. Compared to \textit{NN-LUT}, it achieves improvements of $1.07\%$ and $0.88\%$ for these models.
\vspace{-1mm}
\subsection{Hardware Performance}
In order to further prove the necessity and superiority of INT8 LUT-based $pwl$, we implement the two types of hardware units depicted in Figure (\ref{Figure1}) using Verilog HDL. The area and power dissipation of each LUT-based unit are obtained through synthesis using Synopsys Design Compiler, utilizing TSMC's 28-nm Technology. To ensure a fair comparison, the operating frequency is set to 500MHz for all hardware units. The comparison results are shown in Table \ref{table6} where the FP32 denotes the LUT-based $pwl$ without input quantization representing the methods employed by \textit{NN-LUT} and \textit{RI-LUT}. The results reveal that an 8-entry INT8 LUT-based $pwl$ requires merely 961um$^2$ of area, achieving remarkable reductions of 81.3\% and 81.7\% compared to high-precision FP32 and INT32 units, respectively. In terms of power dissipation, it consumes 0.4mW resulting in substantial savings of 80.2\% in FP32 and 79.3\% in INT32. Moreover, expanding the LUT storage to 16 entries results in an approximate 1.71$\times$ increase in area and 1.95$\times$ in power relative to the 8-entry INT8 configuration. Therefore, a smaller entry size and low-precision are crucial for $pwl$ under integer-only quantization.
\vspace{-1mm}
\section{Conclusion}
In this study, we introduce a new approach to LUT approximation emphasizing quantization awareness and reveal the phenomenon of breakpoint deviation in the integer-only quantization containing large scaling factors. To handle this, we propose a unique genetic quantization-aware $pwl$ algorithm \textit{GQA-LUT}, and an enhancement technique \textit{RM}, which images the FXP conversion as a mutation process. The experimental results demonstrate that \textit{GQA-LUT} with \textit{RM} surpasses the previous state-of-the-art work \textit{NN-LUT}, in both operator-level performance and fine-tuning accuracy. Additionally, the INT8 hardware $pwl$ units integrated into \textit{GQA-LUT} yield significant savings of 81.3\textasciitilde81.7\% in area and 79.3\textasciitilde80.2\% in power dissipation compared to their high-precision counterparts.
\vspace{-1mm}
\section{Acknowledgement}
This work was supported by ACCESS – AI Chip Center for
Emerging Smart Systems, sponsored by InnoHK funding, Hong
Kong.

%%
%% The next two lines define the bibliography style to be used, and
%% the bibliography file.
\bibliographystyle{unsrt}
\bibliography{citation}

\begin{thebibliography}{10}

\bibitem{kenton2019bert}
Jacob~Devlin et~al.
\newblock Bert: Pre-training of deep bidirectional transformers for language understanding.
\newblock In {\em Proceedings of naacL-HLT}, volume~1, page~2, 2019.

\bibitem{liu2021swin}
Ze~Liu~et al.
\newblock Swin transformer: Hierarchical vision transformer using shifted windows.
\newblock In {\em Proceedings of the IEEE/CVF international conference on computer vision}, pages 10012--10022, 2021.

\bibitem{xie2021segformer}
Enze Xie~et al.
\newblock Segformer: Simple and efficient design for semantic segmentation with transformers.
\newblock {\em Advances in Neural Information Processing Systems}, 34:12077--12090, 2021.

\bibitem{Cai_2023_ICCV}
Han Cai~et al.
\newblock Efficientvit: Lightweight multi-scale attention for high-resolution dense prediction.
\newblock In {\em Proceedings of the IEEE/CVF International Conference on Computer Vision (ICCV)}, pages 17302--17313, October 2023.

\bibitem{zhang2023augmented}
Dong Zhang~et al.
\newblock Augmented fcn: rethinking context modeling for semantic segmentation.
\newblock {\em Science China Information Sciences}, 66(4):142105, 2023.

\bibitem{kim2021bert}
Sehoon Kim~et al.
\newblock I-bert: Integer-only bert quantization.
\newblock In {\em International conference on machine learning}, pages 5506--5518. PMLR, 2021.

\bibitem{pmlr-v202-liu23w}
Shih-Yang Liu~et al.
\newblock Oscillation-free quantization for low-bit vision transformers.
\newblock In {\em Proceedings of the 40th International Conference on Machine Learning}, volume 202, pages 21813--21824. PMLR, 23--29 Jul 2023.

\bibitem{liu2023llm}
Shih-yang Liu~et al.
\newblock Llm-fp4: 4-bit floating-point quantized transformers.
\newblock In {\em Proceedings of the 2023 Conference on Empirical Methods in Natural Language Processing}, pages 592--605, 2023.

\bibitem{tu2023multcim}
Fengbin Tu~et al.
\newblock Multcim: Digital computing-in-memory-based multimodal transformer accelerator with attention-token-bit hybrid sparsity.
\newblock {\em IEEE Journal of Solid-State Circuits}, 2023.

\bibitem{stevens2021softermax}
Jacob~R Stevens~et al.
\newblock Softermax: Hardware/software co-design of an efficient softmax for transformers.
\newblock In {\em 2021 58th ACM/IEEE Design Automation Conference (DAC)}, pages 469--474. IEEE, 2021.

\bibitem{yu2022nn}
Joonsang Yu~et al.
\newblock Nn-lut: neural approximation of non-linear operations for efficient transformer inference.
\newblock In {\em 2023 59th ACM/IEEE Design Automation Conference (DAC)}, pages 577--582, 2022.

\bibitem{kim2023range}
Janghyeon Kim~et al.
\newblock Range-invariant approximation of non-linear operations for efficient bert fine-tuning.
\newblock In {\em 2023 60th ACM/IEEE Design Automation Conference (DAC)}, pages 1--6. IEEE, 2023.

\bibitem{huang2022sdq}
Xijie Huang~et al.
\newblock Sdq: Stochastic differentiable quantization with mixed precision.
\newblock In {\em International Conference on Machine Learning}, pages 9295--9309. PMLR, 2022.

\bibitem{10071554}
Xianghong Hu~et al.
\newblock A tiny accelerator for mixed-bit sparse cnn based on efficient fetch method of simo spad.
\newblock {\em IEEE Transactions on Circuits and Systems II: Express Briefs}, 70(8):3079--3083, 2023.

\bibitem{jacob2018quantization}
Benoit Jacob~et al.
\newblock Quantization and training of neural networks for efficient integer-arithmetic-only inference.
\newblock In {\em Proceedings of the IEEE conference on computer vision and pattern recognition}, pages 2704--2713, 2018.

\bibitem{vaswani2017attention}
Ashish Vaswani~et al.
\newblock Attention is all you need.
\newblock {\em Advances in neural information processing systems}, 30, 2017.

\bibitem{han2023flatten}
Dongchen Han~et al.
\newblock Flatten transformer: Vision transformer using focused linear attention.
\newblock In {\em Proceedings of the IEEE/CVF International Conference on Computer Vision}, pages 5961--5971, 2023.

\bibitem{bengio2013estimating}
Yoshua Bengio~et al.
\newblock Estimating or propagating gradients through stochastic neurons for conditional computation.
\newblock {\em arXiv preprint arXiv:1308.3432}, 2013.

\bibitem{DBLP:conf/iclr/EsserMBAM20}
Steven K.~Esser et~al.
\newblock Learned step size quantization.
\newblock In {\em 8th International Conference on Learning Representations, {ICLR} 2020, Addis Ababa, Ethiopia, April 26-30, 2020}, 2020.

\bibitem{holland1992adaptation}
John Holland.
\newblock {\em Adaptation in natural and artificial systems: an introductory analysis with applications to biology, control, and artificial intelligence}.
\newblock MIT press, 1992.

\bibitem{cordts2016cityscapes}
Marius Cordts~et al.
\newblock The cityscapes dataset for semantic urban scene understanding.
\newblock In {\em Proceedings of the IEEE conference on computer vision and pattern recognition}, pages 3213--3223, 2016.

\bibitem{zhang2022graph}
Dong Zhang~et al.
\newblock Graph reasoning transformer for image parsing.
\newblock In {\em Proceedings of the 30th ACM International Conference on Multimedia}, pages 2380--2389, 2022.

\end{thebibliography}
    
%%
%% If your work has an appendix, this is the place to put it.
\appendix
\end{document}